\documentclass{article}

\PassOptionsToPackage{numbers, compress}{natbib}

\usepackage{PRIMEarxiv}
\usepackage{color}
\usepackage{natbib}
\usepackage{wrapfig}

\usepackage{amssymb}
\usepackage[utf8]{inputenc}
\usepackage[T1]{fontenc}
\usepackage{hyperref}
\usepackage{url}
\usepackage{booktabs}
\usepackage{amsfonts}
\usepackage{nicefrac}
\usepackage{microtype}
\usepackage{array}
\usepackage{multicol}
\usepackage{multirow}
\usepackage[utf8]{inputenc}
\usepackage{tabularx}
\usepackage[T1]{fontenc}
\usepackage{graphicx}
\usepackage{enumitem}
\usepackage[table,xcdraw]{xcolor}
\usepackage{amsmath}
\usepackage{fontawesome}
\usepackage{makecell}

\usepackage[utf8]{inputenc}

\definecolor{bb}{rgb}{0.12,0.565,1}
\definecolor{gg}{rgb}{0.2,0.8,0.2}
\definecolor{rr}{rgb}{1,0.85,0.2}

\newif\ifdraft
\drafttrue

\hypersetup{
    colorlinks=true,
}

\newcommand{\ours}[0]{Baichuan-omni\ }
\title{Baichuan-Omni Technical Report}
\author{
    Yadong Li$^{1}$\thanks{Equal Core Contributors. }\enskip\enskip
    Haoze Sun$^{1}$\footnotemark[1]\enskip\enskip
    Mingan Lin$^{1}$\footnotemark[1]\enskip\enskip
    Tianpeng Li$^{1}$\footnotemark[1]\enskip\enskip
    Guosheng Dong$^{1}$\footnotemark[1]\enskip\enskip\\
    Tao Zhang$^{1}$\enskip\enskip 
    Bowen Ding$^{2,3}$\enskip\enskip
    Wei Song$^{2,3}$\enskip\enskip
    Zhenglin Cheng$^{2,3}$\enskip\enskip
    Yuqi Huo$^{1}$\enskip\enskip \\
    Song Chen$^{1}$\enskip\enskip
    Xu Li$^{1}$\enskip\enskip 
    Da Pan$^{1}$\enskip\enskip
    Shusen Zhang$^{1}$\enskip\enskip
    Xin Wu$^{1}$\enskip\enskip
    Zheng Liang$^{1}$\enskip\enskip \\
    Jun Liu$^{1}$\enskip\enskip
    Tao Zhang$^{1}$\enskip\enskip
    Keer Lu$^{1}$\enskip\enskip
    Yaqi Zhao$^{1}$\enskip\enskip
    Yanjun Shen$^{1}$\enskip\enskip
    Fan Yang$^{1}$\enskip\enskip \\
    Kaicheng Yu$^{2}$\enskip\enskip
    Tao Lin$^{2}$\enskip\enskip
    Jianhua Xu$^{1}$\thanks{Corresponding author.}\enskip\enskip
    Zenan Zhou$^{1}\footnotemark[2]$\enskip\enskip
    Weipeng Chen$^{1}$\enskip\enskip \\
    \textsuperscript{1} Baichuan Inc. \enskip
    \textsuperscript{2} Westlake University  \enskip 
    \textsuperscript{3} Zhejiang University \enskip 
    \\
    \texttt{\{xujianhua, zhouzenan\}@baichuan-inc.com}
    \enskip \\
}

\begin{document}

\maketitle

\begin{center}
  \vspace{-3em}
  \faGithub~\url{https://github.com/westlake-baichuan-mllm/bc-omni}
  \vspace{0.75em}
\end{center}

\begin{figure*}[ht]
  \vspace{-2em}
  \centering
  \begin{minipage}[t]{0.497\textwidth}
    \centering
    \includegraphics[height=8.0cm]{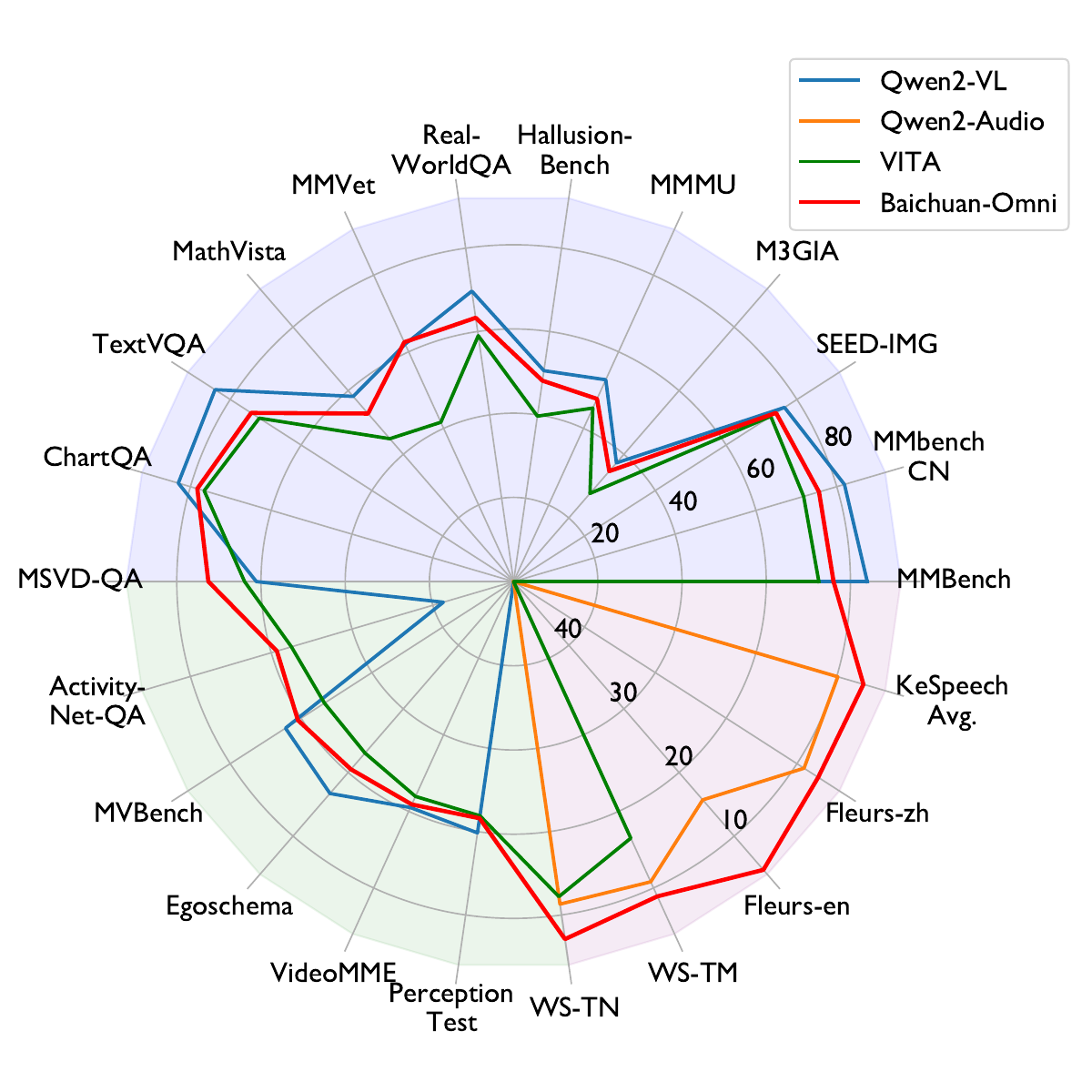}
  \end{minipage}
  \begin{minipage}[t]{0.497\textwidth}
    \centering
    \includegraphics[height=7.5cm]{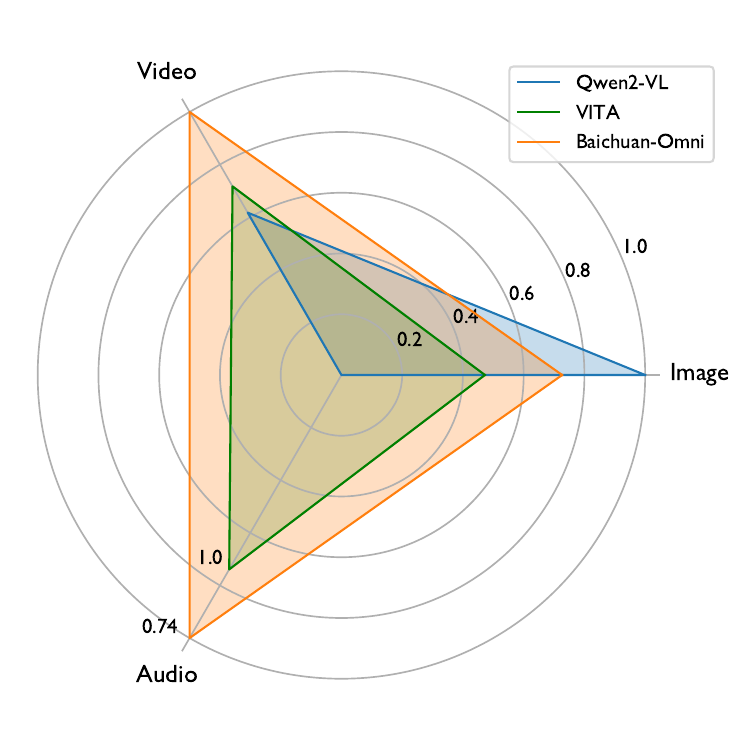}
  \end{minipage}
  \vspace{-1.5em}
  \caption{\textbf{Evaluation across image, video, and audio modalities.} \textbf{(Left)} \ours covers more modalities than Qwen2 VL~\citep{teamQwen2VLSeeWorld2024} and outperforms the current leading omni-modal model, VITA~\citep{fu2024vita}. \textbf{(Right)} Average scores across benchmarks for all modalities. All the scores are normalized by $x_{\text{norm}}=(x-x_{\text{min}}+10)/(x_{\text{max}}-x_{\text{min}}+10)$.}
  \label{fig:lidar}
\end{figure*}

\begin{abstract}

The salient multimodal capabilities and interactive experience of GPT-4o highlight its critical role in practical applications, yet it lacks a high-performing open-source counterpart.  
In this paper, we introduce \textbf{\ours}, the first open-source 7B Multimodal Large Language Model (MLLM) adept at concurrently processing and analyzing modalities of image, video, audio, and text, while delivering an advanced multimodal interactive experience and strong performance. We propose an effective multimodal training schema starting with 7B model and proceeding through two stages of multimodal alignment and multitask fine-tuning across audio, image, video, and text modal. 
This approach equips the language model with the ability to handle visual and audio data effectively. Demonstrating strong performance across various omni-modal and multimodal benchmarks, we aim for this contribution to serve as a competitive baseline for the open-source community in advancing multimodal understanding and real-time interaction.

\end{abstract}

\section{Introduction}

The burgeoning field of artificial intelligence has witnessed a remarkable evolution, especially with the development of Large Language Models (LLMs)~\cite{achiam2023gpt,brown2020language,zhao2023survey} and the subsequent emergence of Multimodal Large Language Models (MLLMs)~\cite{li2024mini,HelloGPT4o,yin2023survey}, signifying a paradigm shift in how machines understand and interact with the world. 
The introduction of MLLMs like GPT-4o~\cite{HelloGPT4o}, characterized by their exceptional multimodal capabilities and enriched interactive experiences, has not only spotlighted the indispensable role of these technologies in real-world applications but also set a new benchmark for what is achievable in terms of human-computer interaction. 

Despite the remarkable progress of MLLMs, current open-source solutions exhibit notable deficiencies, particularly in multimodal capabilities and the quality of user interaction experiences~\cite{fu2024vita}. These shortcomings significantly impede the broader adoption and effectiveness of such models in diverse applications, from natural language processing~\cite{devlin2018bert,radford2019language} to computer vision~\cite{wang2024visionllm,song2023llm} and beyond. 

In response to these challenges, we introduce an omni-modal LLM \textbf{\ours} alongside a multimodal training scheme designed to facilitate advanced multimodal processing and naturalistic user interactions. The architecture of \ours is depicted in \autoref{fig:architecture}. 
The scheme of \ours is built upon three core components:

\begin{figure*}[!ht]
    \centering
    \includegraphics[width=\textwidth]{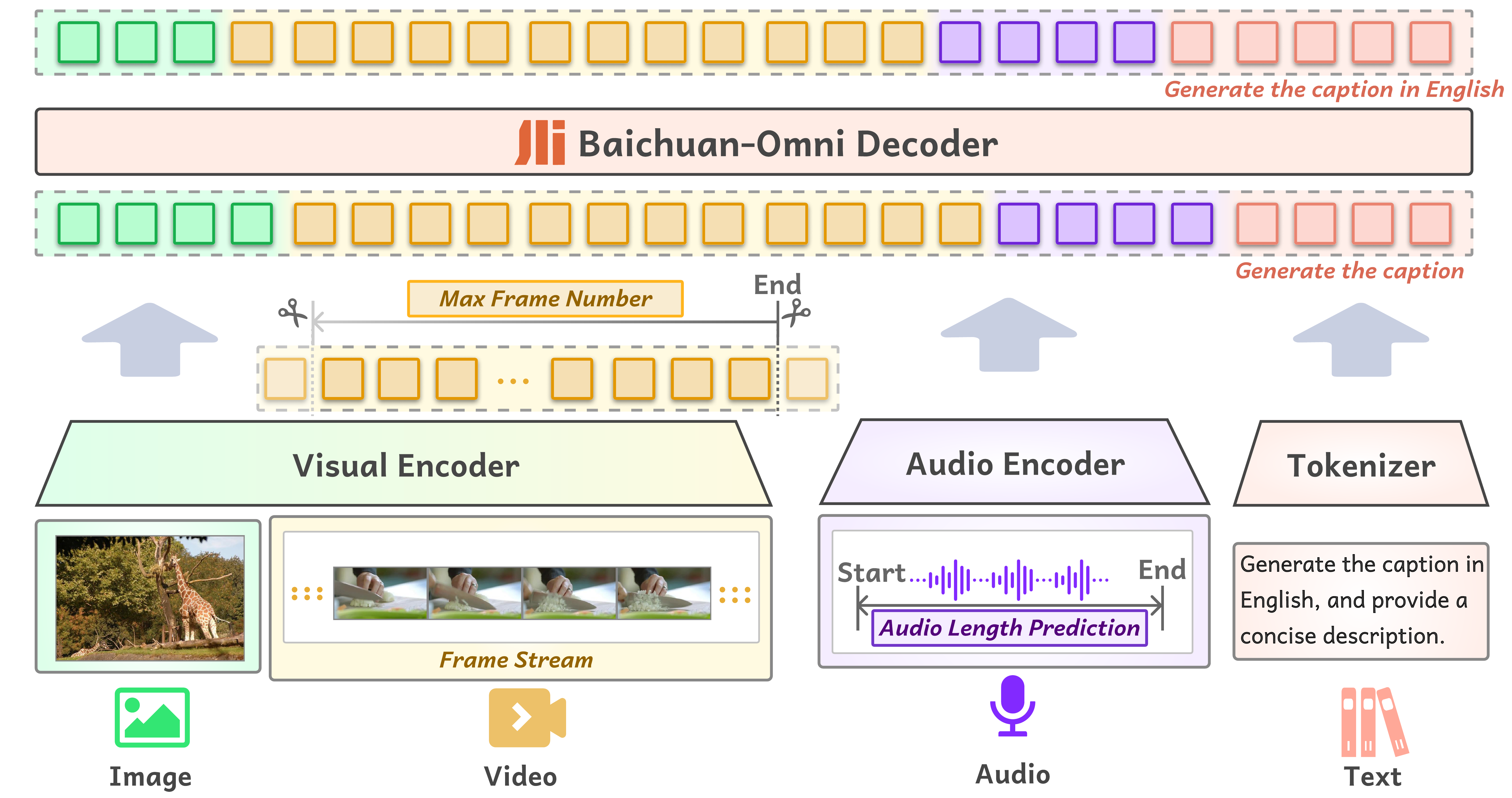}
    \caption{\textbf{Architecture of \ours.} \ours is designed to process both pure text/audio inputs and combinations of video/image with text/audio. In terms of interactivity, the model initially predicts the start and end of audio inputs. During this period, incoming images and videos are encoded into features and fed into the MLLM in a streaming fashion to calculate attention. 
    The audio features are then input into the MLLM for inference following the end of the audio input, facilitating streaming input of audio and video.}
    \label{fig:architecture}
\end{figure*}

\textbf{Omni-Modal Data Construction.} \quad 
We utilize a substantial collection of high-quality, omni-modal data to train \ours with a blend of open-source, synthetic, and internally annotated datasets. In the multimodal alignment pre-training phase, we curate a wide-ranging assortment of training corpora that encompasses image captions, interleaved data, OCR data, and image-text data. For audio alignment, we collect both open-source and in-house datasets for Automatic Speech Recognition (ASR) and Audio Question Answering (AQA). In the realm of video alignment, we acquire video data from both open-source and in-house sources. During the multimodal supervised fine-tuning phase, we compile and synthesize an extensive dataset that covers over 200 tasks and comprises 600,000 instances across pure text, audio, image-text, video-text, and image-audio interaction data.

\textbf{Multimodal Alignment.} \quad 
During the pre-training phase for multimodal alignment, we meticulously align encoders and connectors across various modalities. Initially, we train the vision-language model using a substantial dataset of image-text pairs. This foundational training enables us to harness the visual capabilities developed during the image-text training to further train the video projector. Concurrently, we train the audio-language model utilizing Automatic Speech Recognition (ASR) data. Building upon this robust foundation, we integrate high-quality image, audio, and video data to achieve comprehensive multimodal alignment.

\textbf{Multitask Fine-tuning.} \quad 
For the omni-modal fine-tuning stage, we utilize a multi-task cross-modal interaction training corpus derived from a combination of open-source, synthetic, and internally annotated data. We select data for the final supervised fine-tuning (SFT) phase based on criteria that whether factual knowledge is already learned by the pre-trained model~\cite{gekhman2024doesfinetuningllmsnew}. During this phase, we implement a packing technique to concatenate multiple samples, using the \texttt{cuseq\_len} from \texttt{flash-attention2} for effective sample isolation. 
With this, multiple samples can be packaged into a large batch while ensuring each sample is correctly isolated during the computational process, preventing data confusion between different samples. This accelerates the training process and optimizes memory usage.

The contributions of this paper are summarized below:

\begin{itemize}[leftmargin=*]
    \item %
    We introduce \ours, an open-source, high-performance foundational omni-modal model capable of concurrently processing text, images, videos, and audio inputs. It also provides multilingual support for languages including English and Chinese. Our training framework features a comprehensive pipeline that includes the construction of omni-modal training data, multimodal alignment pre-training, and multimodal supervised fine-tuning, with a particular emphasis on enhancing omni-modal instruction-following capabilities.
    \item %
    We explore early-stage research in natural multimodal human-computer interactions. Our approach initiates with the prediction of audio input boundaries, while simultaneously streaming and encoding incoming visual data into features. These features are then processed by a multimodal large language model (MLLM) for dynamic attention computation. Upon completion of the audio input, the corresponding features are fed into the MLLM for inference, thus facilitating the support for handling audio and video inputs. This integrated approach allows for real-time processing and enhances the interactive capabilities of the system. 
    \item %
    We have made our \ours model, training code, and evaluation scripts publicly available, aiming at fostering progress within the research community. As pioneers in this field, we remain committed to furthering the development of multimodal foundational models and their interactions.
\end{itemize}

\section{Related works}

Recent advancements in Large Language Models (LLMs) have reshaped the AI landscape, paving the way for the emergence of Multimodal Large Language Models (MLLMs).
These advanced models expand AI capabilities beyond text, allowing understanding and generation of content across multiple modalities, including images, audio, and video, signaling a significant leap in AI development.

Open-source MLLMs have demonstrated increasingly powerful capabilities, with efforts from both academia and industry fueling the rapid development of models.
LLMs such as LLaMA~\cite{touvron2023llama,touvron2023llama2}, MAP-Neo~\cite{zhang2024mapneo}, Baichuan~\cite{yang2023baichuan2}, Qwen~\cite{bai2023qwen,yang2024qwen2}, and Mixtral~\cite{jiang2023mistral} are trained on extensive text data, exhibiting strong capacities in natural language comprehension and task resolution through text generation. 
Vision-Language Models (VLMs)~\cite{li2023blip,zhu2023minigpt,zhang2023videollama} have shown promising potential in addressing vision-focused issues, with representative models including LLaVA~\cite{liu2024visual}, DeepSeek-VL~\cite{lu2024deepseek}, the Qwen-VL series~\cite{bai2023qwenvl,teamQwen2VLSeeWorld2024}, InternVL families~\cite{chen2023internvl,chen2024far}, and MiniCPM~\cite{hu2024minicpm}. 
Additionally, Audio-Language Models (ALMs)~\cite{wu2024towards,deshmukh2023pengi,kong2024audio} leverage audio-text pairs to perceive audio signals based on a singular audio encoder. Notable instances of these models encompass Qwen-Audio~\cite{chu2023qwen,chu2024qwen2}, SALMONN~\cite{tang2024salmonn}, SpeechGPT~\cite{zhang2023speechgpt}, etc.

However, compared to proprietary models like GPT-4o~\cite{HelloGPT4o}, open-source models still exhibit substantial gaps in their capabilities for multimodal interactions,
and there is a considerable scarcity of open-source models that effectively facilitate comprehensive multimodal interactions~\cite{fu2024vita}. To address these, we propose \ours, an open-source and capable MLLM which concurrently supports interactions across modalities including audio, image, video, and text.

\section{Training}

\subsection{High-Quality Multimodal Data}
\label{data}
For training an omni-modal model with strong ability, we build an extensive cross-modal dataset with high quality, including text, image-text,
video-text, audio-text, and their interactions. 

\textbf{Image Data.} \quad Image data can be categorized into several types: Caption, Interleaved image-text, OCR data and Chart data~\cite{jain1981image}. From the perspective of sources, it is divided into Open-source data and Synthetic data. Regarding open-source data, we have collected major open-source datasets, including PIN-14M~\citep{2406.13923}, MINT-1T~\citep{awadalla2024mint1t}, LAION-5B~\citep{schuhmann2022laion5bopenlargescaledataset}, OBELIC~\citep{laurencon2023obelics}, etc. for Stage I training of Image-language branch (Detailed introduction in Section \ref{sec:image-lang-branch}), and Cauldron~\citep{laurençon2024matters}, Monkey~\citep{li2023monkey}, ArxivQA~\citep{li-etal-2024-multimodal-arxiv}, TGDoc~\citep{wang2023towards}, MM-Self-Instruct (Train split)~\citep{zhang2024multimodal}, MMTab~\citep{zheng2024multimodaltableunderstanding}, etc. for Stage II/III training of Image-language branch. These publicly available open-source datasets are subjected to a series of processing steps and careful sampling techniques within our data pipeline.

\par As for synthetic data, the purpose is to obtain higher quality data to enhance the performance of our models. One part is derived from books and papers, which are parsed to generate Interleaved image-text, OCR data and Chart data. It is highly complete and specialized, making it high-quality and knowledge intensive data. Another part involves training dedicated models to produce image captions. These captions describe the content of the images in detail from different perspective, belonging to high-quality caption data.

\textbf{Video Data.} \quad Video dataset comprises a diverse array of publicly available resources, encompassing multiple tasks such as video classification, action recognition, and temporal localization. The video-text sources can be categorized into two main types: question-answering (QA) data and caption data.

\par For QA data, we incorporate: NExTVideo, introduced in LLaVA-NExT~\citep{zhang2024llavanext-video} and ActivityNet-QA (Train split)~\citep{yu2019activitynet}. Our caption data sources include ShareGPT4Video~\citep{chen2024sharegpt4videoimprovingvideounderstanding}, a large-scale dataset that leverages GPT-4 to generate rich, contextual captions for videos, and WebVid~\citep{bain2021frozen}. To further enrich our dataset, we have employed GPT-4o to generate diverse captions for videos collected from YouTube.

\par The sampling ratio for each dataset within our compilation is carefully determined based on the relative sizes of these datasets. This strategic approach ensures a balanced representation of various video types, tasks, and domains in our final dataset.

\textbf{Audio Data.} \quad Considering the diversity of audio data, we extract audio from various media modalities, which includes different recording environments, languages, accents, and speakers. Guided by the principles in previous work~\citep{radford2022robustspeechrecognitionlargescale}, we posit that the variation in audio quality contributes to a robust speech understanding capability. To facilitate a more sophisticated classification and filtering procedure, we implemented a data processing pipeline comprising speaker voice recording, dialect recognition, accent recognition, sound effect detection, and quality assessment.

To enhance the quality of audio-text pairs derived from the dataset, we utilized an in-house ASR system along with several open-source models~\cite{radford2022robustspeechrecognitionlargescale,gao2022paraformer,speechteam2024funaudiollm} to generate multiple transcript versions. These generated data are then refined through a model ensemble strategy for effective text filtering and error correction.

\textbf{Text Data.} \quad In handling text corpus, we collected data from various domains such as web pages, books, academic papers, code, etc.. Following the data processing protocols proposed in previous works~\citep{dong2024baichuanseed,lu2024datasculpt}, we implemented a selection process to enhance the diversity and quality of the dataset. The diversity criterion ensures broad coverage of topics and linguistic styles in the training corpus, accommodating various applications. High-quality processing removes redundancy and noise from the text data, increasing knowledge density.

\textbf{Cross-Modal Interaction Data.} \quad 
To enhance the cross-modal interaction capabilities of our model, we synthesized a collection of visual-audio-text cross-modal interaction data, including both image-audio-text and video-audio-text datasets. For the image-text data, we segmented the textual data into a 1:3 ratio, converting the initial quarter of text into audio descriptions using text-to-speech (TTS) technology. Our audio encompasses 44 different timbres, ensuring a diversity of vocal tones. This setup is complemented by task prompts such as ``Please listen to the following audio describing the content of the image. Your task is to supplement more information by integrating the image after listening'', aiming to predict the remaining three-quarters of the textual description. For the video-text data, we directly extracted the audio from the videos to serve as the cross-modal audio component.

\begin{figure*}[!ht]
    \centering
    \includegraphics[width=\textwidth]{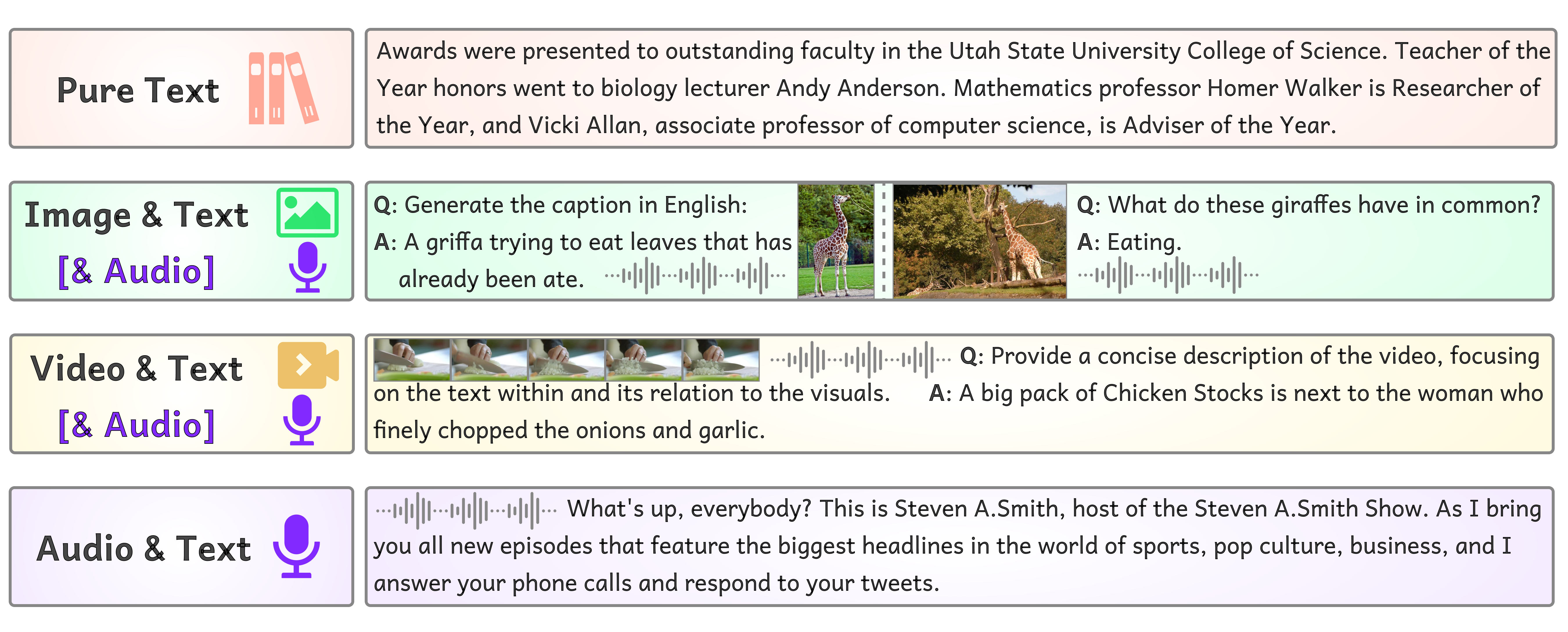}
    \caption{\textbf{Data illustration of \ours.} We build an extensive cross-modal dataset, including text, image-text, video-text, audio-text, and their interactions. Our collection also features integrated image-audio-text and video-audio-text data.}
    \label{fig:data_example}
\end{figure*}

\subsection{Multimodal Alignment Pre-training}

\begin{figure*}[!ht]
    \centering
    \includegraphics[width=\textwidth]{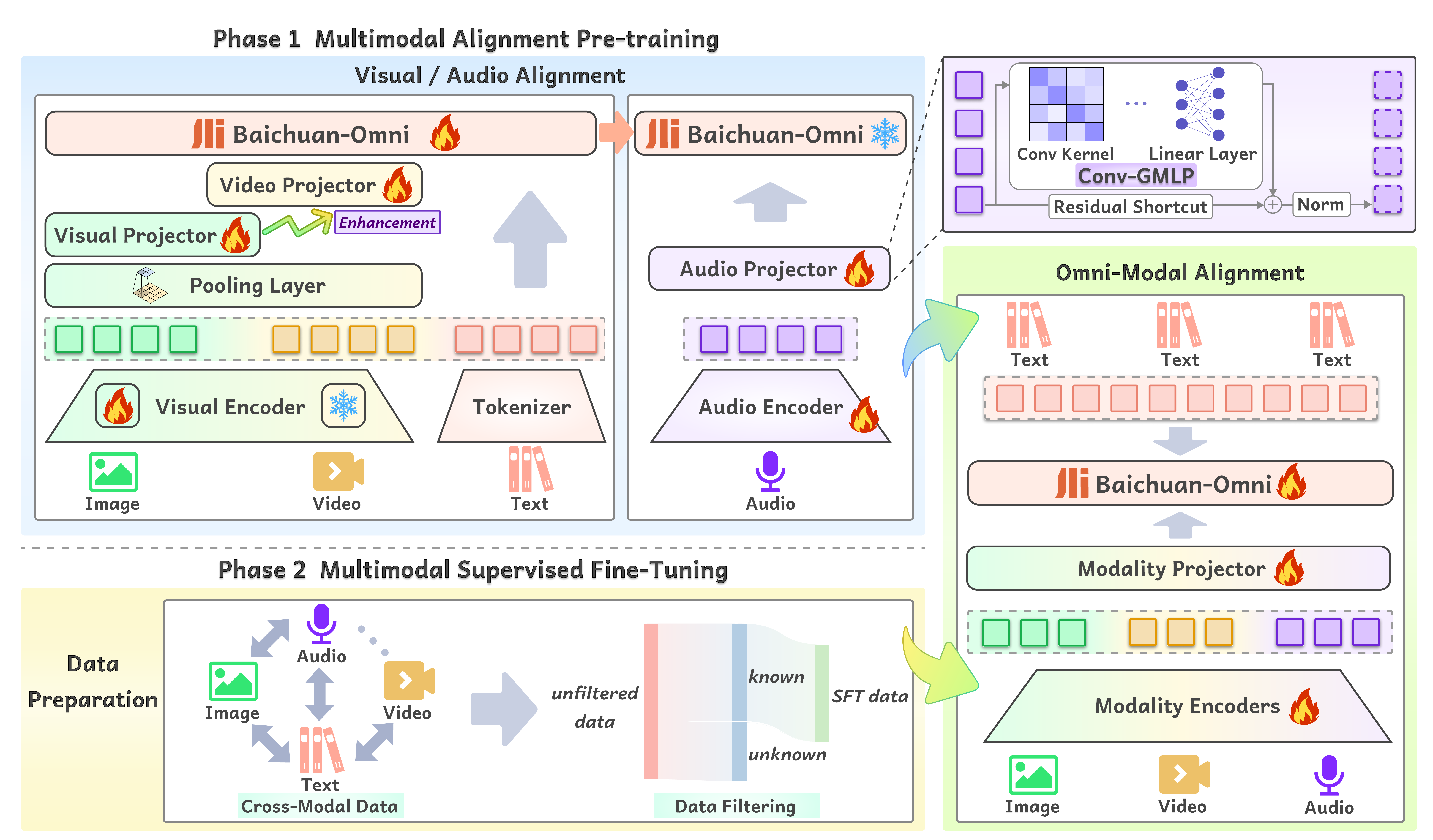}
    \caption{\textbf{Training Pipeline of \ours}. During the pretraining phase, we initially train a vision-language model using extensive image-text data, followed by training an audio-language model with ASR data. Subsequently, we integrate high-quality images, audio, and video data for comprehensive multimodal alignment. In the fine-tuning phase, we synthesize a subset of cross-modal interaction data to blend with existing high-quality datasets. From this enriched dataset, we select a subset of data that the model is already capable of handling and proceed with multimodal multitask fine-tuning. This process aims to enhance the model's adherence to omni-modal instructions.}
    \label{fig:pipeline}
\end{figure*}

\par In this section, we will further illustrate the pre-training and alignment processes for the Image-Language, Video-Language, and Audio-Language branches.

\subsubsection{Image-Language Branch}
\label{sec:image-lang-branch}

We utilize Siglip-384px~\cite{zhai2023sigmoid} as the visual encoder, which processes a 384$\times$384 image input and generates 182 tokens through a visual projector composed of a two-layer MLP and a 2$\times$2 convolution layer serving as the pooling layer.
To scale the input to arbitrary resolutions while preserving the intricate details of high-resolution images, we adopt AnyRes~\cite{liu2024llavanext}, which splits the image into grids and concatenates the features of a down-sampled image to provide global context. The training of our image-language branch is divided into three stages.
\begin{itemize}[leftmargin=*]
    \item \textbf{Stage I:} In the first stage, we train the visual projector to establish the initial alignment between image representations and text through image captioning task. During this phase, we freeze the LLM and the visual encoder, only training the visual projector with a learning rate of $1e-3$.
    \item \textbf{Stage II:} In the second stage, we freeze the LLM and train both the projector and visual encoder with a smaller learning rate of $1e-5$. In addition to general VQA tasks, we specifically synthesized 130k high-quality QA data for OCR and charts to enhance the model's abstract visual understanding. We also introduced interleaved data and image caption data in this stage, which help maintain and promote better alignment between image and text representations, mitigating shifts caused by changes in the image feature space after unfreezing the visual encoder.
    \item \textbf{Stage III:} 
    Based on the second stage, we unfreeze the LLM and continue updating the parameters of all model components with a learning rate of $1e-5$ to further enhance visual-language performance. In addition to VQA and image-caption pairs, we also introduce interleaved data and pure text data in this stage to better maintain the original capabilities of the LLM.
\end{itemize}

\subsubsection{Video-Language Branch}

\par Based on the visual capabilities acquired from the pre-training of the Image-Language Branch, we proceed to train the video projector using a frozen vision encoder (Siglip-384px, the same as that used in the Image-Language Branch) alongside an LLM (Large Language Model) backbone. This training process employs a low learning rate of $4e-6$ to refine the alignment with the language modality.

\par During the training phase, the input video frames are sampled at a rate of 1 frame per second, with a maximum of 48 frames per video. Each input frame is resized to a maximum resolution of 384$\times$768 pixels to maintain optimal quality and detail. Furthermore, a 2$\times$2 convolution layer is applied prior to the video projector. This convolutional step serves to regulate the length of the video token sequence, ensuring a minimum of 182 tokens and a maximum of 546 tokens. This thoughtful configuration strikes a balance between performance and efficiency, facilitating effective model training while managing the computational load.

\par Rather than immediately proceeding with the pre-training of the Video-Language Branch using only pure video-text pairs, we have opted for a more nuanced two-stage approach. Initially, we leverage image-text pre-training data to strengthen the model's visual understanding capabilities. After establishing a robust foundation, we incrementally integrate mixed image-text pairs and video-text pairs into the training regimen. This strategy has proven to yield superior results. By gradually enhancing the model's visual competence, we provide valuable guidance for the video pre-training pipeline, allowing the model to better understand and integrate the complexities of video data in conjunction with language. This methodology underscores the importance of a comprehensive training strategy that incorporates diverse data modalities for improved alignment and performance.

\subsubsection{Audio-Language Branch}\label{sec: audio_branch_training}
\par The Audio-Language branch extends an LLM pre-trained on visual and video data by incorporating an audio encoder from the Whisper-large-v3 model~\cite{radford2022robustspeechrecognitionlargescale} and a newly introduced audio projector.

The audio encoder processes the audio signal (30s, 128 mel-spectrum) into an audio representation in a 1280-channel feature space, while the audio projector (typically a linear projector~\citep{chu2024qwen2} or MLP) maps that to the embedding space of LLM. Prior to projection, a pooling operation with a stride of $n$ is traditionally used to down-sample the audio representation into fewer tokens (i.e., frames) for the downstream LLM. However, when we aggressively reduce the number of audio tokens, this simple pooling approach leads to a loss of audio information. In our approach, we replace the pooling with \textbf{Conv}olutional-\textbf{G}ated \textbf{MLP} (Conv-GMLP), leveraging convolution layers for down-sampling to preserve more audio information.

\begin{figure*}[!ht]
    \centering
    \includegraphics[width=\textwidth]{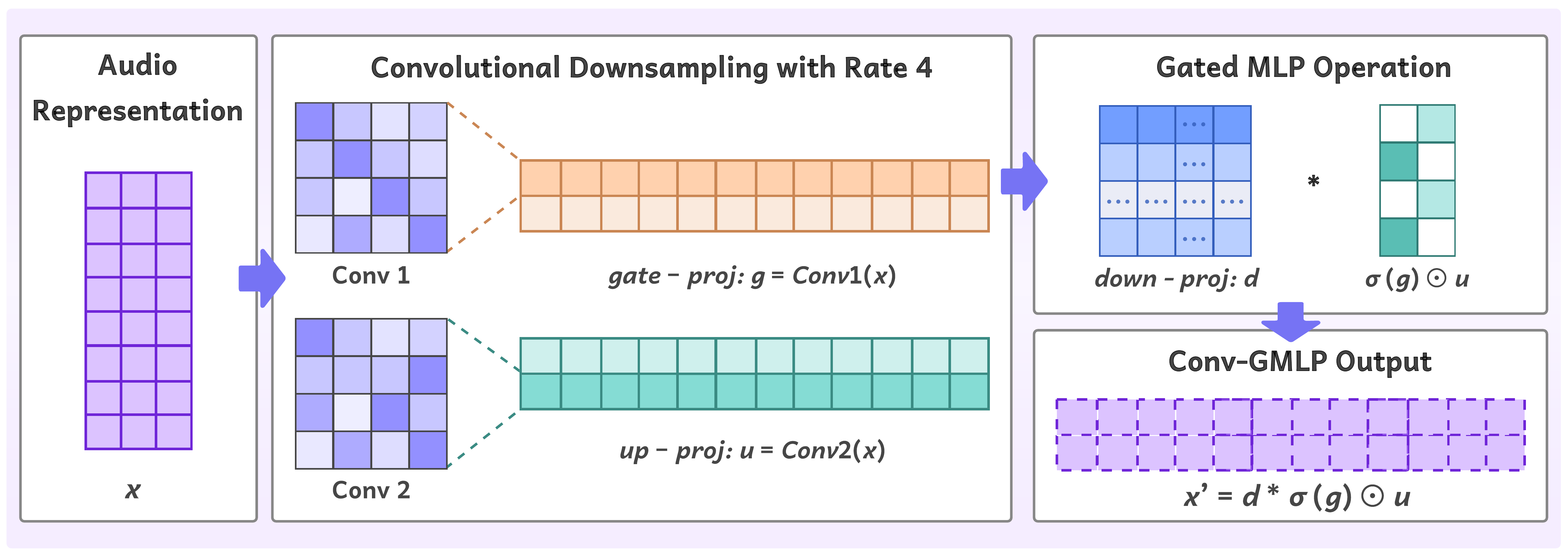}
    \caption{\textbf{Illustration of Conv-GMLP.} Conv-GMLP down-sampling is applied to the audio representation. With a down-sampling rate of 4, the output sequence length is reduced to a quarter of the input, while the number of feature channels increases fourfold.}
    \label{fig:Conv-GMLP}
\end{figure*}

Figure~\ref{fig:Conv-GMLP} illustrates the Conv-GMLP architecture, which functions similarly to a gated MLP~\citep{liu2021payattentionmlps} but replaces linear layers with convolutional ones. Each of the two convolutional layers reduces the sequence length of the audio representation by a factor of \textit{n}, while proportionally expanding the feature space. In our projector, a residual shortcut is along with Conv-GMLP, enabling more efficient gradient back-propagation. Results in Section~\ref{sec: audio-ablation} demonstrate strong robustness in audio performance when setting the down-sampling rate\footnote{The \textit{down-sampling rate} is named as \texttt{avg\_pooler} in our code configuration.} \textit{n} aggressively.

\par During training, the LLM remains frozen, and only the audio encoder and projector are trained using long audio-text sequences (up to 4K tokens). A cosine learning rate scheduler is employed to enhance performance.

\subsubsection{Image-Video-Audio Omni-Alignment}\label{sec: omni-align}

The right part of~\autoref{fig:pipeline} illustrates the `Omni-Alignment' stage, which follows the individual training of the Image-Language, Video-Language, and Audio-Language branches. During this stage, all modules are trained together on a mixture of high-quality image-text, video-text, and audio-text pairs to develop comprehensive multimodal understanding.

\subsection{Multimodal Supervised Fine-Tuning}
\label{sft_data}

\par In this section, we describe the multimodal supervised fine-tuning process aimed at improving the model's ability to follow complex, multimodal instructions across various tasks. We leveraged a diverse set of open-source, synthetic, and internally annotated data, covering over 200 distinct tasks and comprising approximately 600K pairs across text, audio, image-text, video-text, and image-audio modalities.

\paragraph{Text-only Data.} The text-only data covers a broad range of tasks, including knowledge-based question answering, mathematics, logical reasoning, code generation, text creation, information processing, persona-based tasks, and safety-related data. To further strengthen the model's ability to handle complex, multi-step tasks, we included specialized datasets that feature intricate instructions, some of which contain a system message designed to structure more elaborate scenarios.

\paragraph{Image Understanding Data.}
For tasks involving image understanding, we primarily utilized the vFLAN dataset~\citep{chen2024allava}, focusing on its instruction-following data. Given the quality issues present in some of the samples, we employed a loss-based filtering method to clean the dataset: 
\begin{enumerate}[leftmargin=*]
    \item We computed the loss for all vFLAN English instruction samples using the pretrained model and fit the resulting values to a Gaussian distribution.
    \item Samples are removed if their loss values fell outside the range of $ \mu \pm \sigma $.
    \begin{enumerate}
        \item Samples with $ loss < \mu - \sigma $ typically included trivial issues, such as cases where the prompt and response content are nearly identical.
        \item Samples with $ loss > \mu + \sigma $ often had significant problems, such as reversed prompt-response pairs or hallucinations in the responses.
    \end{enumerate}
\end{enumerate}

A subset of the cleaned vFLAN instruction data is then translated into Chinese, followed by manual re-annotation to ensure high-quality alignment. Alongside vFLAN, we incorporated several other open-source datasets, including synthdog-en/zh~\citep{kim2022ocr}, handwritten OCR, street view OCR, reference grounding and grounded captioning  duality tasks, and ImageInWords~\citep{garg2024imageinwords}. Most of these datasets are translated into Chinese. For ImageInWords, we ensured that if an image contained a recognizable entity, the corresponding caption explicitly referenced that entity by name (e.g., identifying a Samoyed dog by breed rather than simply labeling it as ``dog'').

Although vFLAN covers 191 tasks, we found that it lacked variety in instruction types. To address this, we sampled data from our textual SFT dataset and rendered some of the prompts as images to increase the diversity of image-based instructions. Additionally, to enhance the model’s mathematical reasoning with images, we used the method from ~\citep{zhou2024mathscapeevaluatingmllmsmultimodal} to generate a large dataset of multimodal math problems involving images.

In experiments, we found that adding too much external world knowledge that the model inherently did not know resulted in diminishing performance returns. To mitigate this, we adopted the filtering method from ~\citep{gekhman2024doesfinetuningllmsnew} to exclude unknown data from the constucted SFT dataset.

\paragraph{Video Understanding Data.} The video-text data is primarily sourced from the VideoInstruct100K dataset~\citep{maaz2023videochatgpt}. While each video in the dataset includes multiple instructions, the instructions tend to be relatively homogeneous, often focusing on simple video descriptions. To enhance the diversity of video-based tasks, we applied semantic deduplication to the instructions for each video and translated the dataset into Chinese, enriching the variety of video-based tasks for the model.

\paragraph{Audio Understanding Data.} Most of the audio data is generated using TTS \footnote{TTS tool: \url{https://github.com/2noise/ChatTTS}}, with prompts derived from text-only, image-text, and video-text datasets. To ensure the quality of the synthesized audio, we transcribed the generated audio using an ASR model and compared the transcriptions with the original prompts. Only those audio samples with accurate transcriptions are retained as final audio prompts. To further enrich the audio data, we included human-recorded audio samples that captured various dialects, accents, and background noises.

In addition to the general QA tasks, we also constructed a specific ASR dataset sourced from open-source data and internal logs. To improve training efficiency, we filtered out easily recognizable samples, focusing instead on more challenging audio data for supervised fine-tuning.

\section{Experiment}

\subsection{Language Performance}
\subsubsection{Evaluation Benchmarks}

We perform evaluations on 4 comprehensive benchmarks, including MMLU~\cite{hendryckstest2021}, CMMLU~\cite{li2023cmmlu}, AGIEval~\cite{zhong2023agieval} and C-Eval~\cite{huang2024c}. 
MMLU includes 57 unique tasks consisting of multiple-choice questions across various fields of knowledge, including the humanities, social sciences, and hard sciences. 
CMMLU represents an extensive assessment framework tailored to assess the sophisticated knowledge and reasoning capabilities of LLMs within the context of Chinese language and culture. 
AGIEval is a human-centric benchmark crafted to evaluate the foundational models' general cognitive and problem-solving abilities, based on official, public, and qualification tests designed for human participants. 
C-EVAL provides the comprehensive Chinese evaluation suite designed to evaluate the advanced knowledge and reasoning skills of LLMs within a Chinese context, encompassing 13,948 multiple-choice questions across 52 varied disciplines, from humanities to science and engineering. 
We conduct all the evaluations with a zero-shot measurement. 

\subsubsection{Major Performance}

We compare \ours with state-of-the-art proprietary multimodal models such as Gemini 1.5 Pro~\citep{reid2024gemini}, and GPT-4o~\citep{HelloGPT4o}, as well as a series of competitive open-source LLMs and MLLMs such as VITA~\citep{fu2024vita}, MAP-Neo~\cite{zhang2024mapneo}, Qwen1.5-Chat~\cite{bai2023qwen}, Llama3-Instruct~\cite{llama3modelcard} and OLMo~\cite{groeneveld2024olmo}. 
We list major results on comprehensive benchmarks in \autoref{tab:language_benchmark1}. 

\begin{table}[!ht]
    \caption{\textbf{Major results on comprehensive benchmarks.} $*$: Officially reported results. $\diamondsuit$: Retrieved results from official leaderboard or recent papers. The rest unlabeled results are reproduced by ourselves.
    }
    \label{tab:language_benchmark1}
    \centering
    \begin{tabular}{@{}ccccc@{}}
        \toprule
        \multicolumn{1}{c|}{}  & \multicolumn{4}{c}{\textbf{Comprehensive Tasks}} \\ \cmidrule(l){2-5} 
        \multicolumn{1}{c|}{\multirow{-2}{*}{\textbf{Model}}} &
          \begin{tabular}[c]{@{}c@{}}MMLU\\ (Acc.)\end{tabular} &
          \begin{tabular}[c]{@{}c@{}}CMMLU\\ (Acc.)\end{tabular} &
          \begin{tabular}[c]{@{}c@{}}AGIEval\\ (Acc.)\end{tabular} &
          \begin{tabular}[c]{@{}c@{}}C-Eval\\ (Acc.)\end{tabular} \\ 
          \midrule
        \multicolumn{5}{c}{\cellcolor[HTML]{EFEFEF}\textit{Proprietary Models}} \\
        \midrule
        \multicolumn{1}{c|}{GPT 4o} & 88.0$^\diamondsuit$ & 78.3$^\diamondsuit$ & 62.3$^\diamondsuit$ & 86.0$^\diamondsuit$ \\
        \midrule
        \multicolumn{5}{c}{\cellcolor[HTML]{EFEFEF}\textit{Open-source Models (Pure text)}} \\
        \midrule
        \multicolumn{1}{c|}{MAP-Neo (7B)} & 58.2 & 55.1 & 33.9 & 57.5 \\
        \multicolumn{1}{c|}{Qwen1.5-Chat (7B)} & 61.5 & 68.0 & 39.3 & 68.8 \\
        \multicolumn{1}{c|}{Llama3-Instruct (8B)} & 67.1 & 51.7 & 38.4 & 50.7 \\
        \multicolumn{1}{c|}{OLMo (7B)} & 28.4 & 25.6 & 19.9 & 27.3 \\
        \midrule
        \multicolumn{5}{c}{\cellcolor[HTML]{EFEFEF}\textit{Open-source Models (Omni-modal)}}          \\ \midrule
        \multicolumn{1}{c|}{VITA (8x7B)} & \textbf{71.0}$^*$ & 46.6 & 46.2$^*$ & 56.7$^*$ \\
        \multicolumn{1}{c|}{\textbf{\ours (7B)}} & 65.3 & \textbf{72.2} & \textbf{47.7} & \textbf{68.9} \\
        \bottomrule
    \end{tabular}
\end{table}

As shown in \autoref{tab:language_benchmark1}, our \ours significantly outperforms open-source, general pure-text LLMs in comprehensive benchmarks.  Compared to the open-source multimodal model VITA, \ours demonstrates a substantial advantage in Chinese benchmarks, such as CMMLU (\textbf{72.2\%} v.s 46.6\%) and C-Eval (\textbf{68.9\%} v.s 56.7\%), and slightly surpasses VITA in AGIEval (\textbf{47.7\%} v.s 46.2\%).

\subsection{Image Understanding}
\subsubsection{Evaluation Benchmarks}

We evaluate \ours on 13 representative vision-language benchmarks, including MMBench-EN, MMBench-CN~\citep{liu2023mmbench}, M3GIA~\citep{song2024m3gia}, SEEDBench~\citep{li2023seed}, RealWorldQA~\citep{Grok-1.5-Vision-Preview}, MMMU~\citep{yue2023mmmu}, MathVista~\citep{lu2023mathvista}, MME~\citep{fu2024mme}, MMVet~\citep{yu2023mm}, TextVQA~\citep{singh2019textvqa}, OCRBench~\citep{liu2024ocrbench}, ChartQA~\citep{masry2022chartqa}, and HallusionBench~\citep{guan2024hallusionbench}. 
To ensure reproducible evaluation results, we use VLMEvalKit~\citep{duan2024vlmevalkit} uniformly for all evaluations. All evaluations are conducted in a zero-shot manner, adhering to the original setup of the models to ensure fair and consistent comparisons across all models and benchmarks.

\subsubsection{Major Performance}

We compare \ours with state-of-the-art proprietary multimodal models such as Gemini 1.5 Pro~\citep{reid2024gemini}, and GPT-4o~\citep{HelloGPT4o}, as well as a series of competitive open-source multimodal models such as VITA~\citep{fu2024vita} and Qwen2-VL~\citep{teamQwen2VLSeeWorld2024}.
We list major results on VQA (Visual Question Answering) benchmarks and results on MCQ (Multi-choice \& Yes-or-No Question) benchmarks in \autoref{tab:image-perf-mcq} and \autoref{tab:image-perf-vqa}.

\begin{table}[!ht]
    \caption{\textbf{Major Results on Multi-choice benchmarks and Yes-or-No benchmarks.} $*$: Officially reported results. $\diamondsuit$: Retrieved results from official leaderboard or recent papers. The rest unlabeled results are reproduced by ourselves.}
    \label{tab:image-perf-mcq}
    \resizebox{\textwidth}{!}{
    \centering
    \begin{tabular}{@{}cccccccc@{}}
        \toprule
        \multicolumn{1}{c|}{}    & \multicolumn{6}{c}{\textbf{Multi-choice \& Yes-or-No Question}} \\ 
        \cmidrule(l){2-8} 
        \multicolumn{1}{c|}{\multirow{-2}{*}{\textbf{Model}}} &
          \begin{tabular}[c]{@{}c@{}}MMBench\\ (Acc.)\end{tabular} &
          \begin{tabular}[c]{@{}c@{}}MMbench-CN\\ (Acc.)\end{tabular} &
          \begin{tabular}[c]{@{}c@{}}M3GIA\\ (Acc.)\end{tabular} &
          \begin{tabular}[c]{@{}c@{}}SEED-IMG\\ (Acc.)\end{tabular} &
          \begin{tabular}[c]{@{}c@{}}MME\\ (Score)\end{tabular} &
          \begin{tabular}[c]{@{}c@{}}MMMU (val)\\ (Acc.)\end{tabular} &
          \begin{tabular}[c]{@{}c@{}}HallusionBench\\ (Acc.)\end{tabular} \\ 
          \midrule
        \multicolumn{8}{c}{\cellcolor[HTML]{EFEFEF}\textit{Proprietary Models}} \\
          \midrule
          \multicolumn{1}{c|}{GPT-4o}             & 83.4$^\diamondsuit$  & 82.1$^\diamondsuit$  & 59.8$^\diamondsuit$  & -  & 2328.7$^\diamondsuit$  & 69.1$^\diamondsuit$  & 55.0$^\diamondsuit$ \\
          \multicolumn{1}{c|}{GPT-4o-mini}        & -  & -  & -  & -  & 2003.4$^\diamondsuit$  & 60.0$^\diamondsuit$  & 46.1$^\diamondsuit$ \\
          \midrule
        \multicolumn{8}{c}{\cellcolor[HTML]{EFEFEF}\textit{Open-source Models (Vision-language)}} \\
          \midrule
          \multicolumn{1}{c|}{Qwen2 VL (7B)}              & 86.4    & 81.9    & 37.3    & 76.5    & 2326.8$^*$    & 52.7       & 50.6$^*$   \\
          \multicolumn{1}{c|}{MiniCPM-Llama3-V 2.5 (8B)}  & 76.7    & 73.3    & 30.3    & 72.4    & 2024.6$^*$    & 45.8$^*$    & 42.5    \\
          \midrule
          \multicolumn{8}{c}{\cellcolor[HTML]{EFEFEF}\textit{Open-source Models (Omni-modal)}}          \\ \midrule
          \multicolumn{1}{c|}{VITA (8x7B)}                & 74.7    & 71.4    & 27.7    & 72.6    & \textbf{2189.1}       & 45.3       & 39.7$^*$   \\
          \multicolumn{1}{c|}{\textbf{\ours (7B)}}                 & \textbf{76.2}    & \textbf{74.9}    & \textbf{34.7}& \textbf{74.1}    & 2186.9       & \textbf{47.3}       & \textbf{47.8}    \\
        \bottomrule
    \end{tabular}}
\end{table}

\begin{table}[!ht]
    \caption{\textbf{Major Results on VQA benchmarks.} $*$: Officially reported results. $\diamondsuit$: Retrieved results from official leaderboard or recent papers. The rest unlabeled results are reproduced by ourselves.}
    \label{tab:image-perf-vqa}
    \resizebox{\textwidth}{!}{
    \centering
    \begin{tabular}{@{}ccccccc@{}}
        \toprule
        \multicolumn{1}{c|}{}    & \multicolumn{6}{c}{\textbf{Visual Question Answering}} \\ 
        \cmidrule(l){2-7}
        \multicolumn{1}{c|}{\multirow{-2}{*}{\textbf{Model}}} &
          \begin{tabular}[c]{@{}c@{}}RealWorldQA\\ (Acc.)\end{tabular} &
          \begin{tabular}[c]{@{}c@{}}MMVet\\ (Acc.)\end{tabular} &
          \begin{tabular}[c]{@{}c@{}}MathVista-mini\\ (Acc.)\end{tabular} &
          \begin{tabular}[c]{@{}c@{}}TextVQA (val)\\ (Acc.)\end{tabular} &
          \begin{tabular}[c]{@{}c@{}}ChartQA\\ (Acc.)\end{tabular} &
          \begin{tabular}[c]{@{}c@{}}OCRBench\\ (Acc.)\end{tabular} \\ 
          \midrule
        \multicolumn{7}{c}{\cellcolor[HTML]{EFEFEF}\textit{Proprietary Models}} \\
          \midrule
          \multicolumn{1}{c|}{GPT-4o}             & 75.4$^\diamondsuit$  & 69.1$^\diamondsuit$  & 63.8$^\diamondsuit$  & -   & 85.7$^\diamondsuit$  & 73.6$^\diamondsuit$ \\
          \multicolumn{1}{c|}{GPT-4o-mini}        & 67.1$^\diamondsuit$  & 66.9$^\diamondsuit$  & 52.4$^\diamondsuit$  &  -  & -     & 78.5$^\diamondsuit$ \\
          \midrule
        \multicolumn{7}{c}{\cellcolor[HTML]{EFEFEF}\textit{Open-source Models (Vision-language)}} \\
          \midrule
          \multicolumn{1}{c|}{Qwen2 VL (7B)}              & 69.7   & 62.0$^*$  & 58.2$^*$  & 84.3$^*$  & 83.0$^*$  & 84.5$^*$    \\
          \multicolumn{1}{c|}{MiniCPM-Llama3-V 2.5 (8B)}  & 63.5   & 52.0     & 54.3$^*$  & 76.6     & 72.0     & 72.5   \\
          \midrule
          \multicolumn{7}{c}{\cellcolor[HTML]{EFEFEF}\textit{Open-source Models (Omni-modal)}}          \\ \midrule
          \multicolumn{1}{c|}{VITA (8x7B)}                & 59.0   & 41.6$^*$  & 44.9$^*$  & 71.8     & 76.6     & 68.5$^*$    \\
          \multicolumn{1}{c|}{\textbf{\ours (7B)}}                 & \textbf{62.6}   & \textbf{65.4}     & \textbf{51.9}     & \textbf{74.3}     & \textbf{79.6}     & \textbf{70.0}    \\
        \bottomrule
    \end{tabular}}
\end{table}

As shown in \autoref{tab:image-perf-mcq} and \autoref{tab:image-perf-vqa}, our \ours comprehensively outperformed VITA-8*7b~\citep{fu2024vita}, which has 12B activated parameters, in multiple visual tasks, both on VQA benchmarks and MCQ benchmarks.
Besides, we also demonstrates competitive performance comparable to, or even better than, open-source image-specialized models like MiniCPM-Llama3-V 2.5~\citep{yao2024minicpm}. Specifically, \ours outperformed MiniCPM-Llama3-V 2.5 on most VQA tasks, including MMBench-CN, SEED-IMG, MME, HallusionBnech and MMMU which requres expert-level perception and reasoning.
However, despite the advantage of incorporating an additional audio modality compared to Qwen2-VL \citep{teamQwen2VLSeeWorld2024}, the performance gap between our model and Qwen2-VL in image tasks remains evident. Furthermore, it is worth noting that, beyond Qwen2-VL, the stark divide between open-source and closed-source models remains substantial.

\subsection{Video Understanding}
\subsubsection{Evaluation Benchmarks}

\par We perform a thorough evaluation on general video understanding tasks (General VQA) and open-ended video question answering (Open-ended VQA) to comprehensively assess the video understanding capabilities of \ours.

\par For general video understanding tasks, we select Perception-Test~\citep{puatruaucean2023perception}, MVBench~\citep{li2024mvbench}, VideoMME~\citep{fu2024video}, and EgoSchema~\citep{mangalam2023egoschema} for long-form video-language understanding. We report top-1 accuracy for all benchmarks. For VideoMME, we report the results under the setting of "w/o subs". For open-ended video question answering part, we choose ActivityNet-QA~\citep{yu2019activitynet} and MSVD-QA~\citep{xu2017video} as evaluation benchmarks. Following previous work~\citep{maaz2023videochatgpt}, we utilize GPT to assess the quality of the response snippets. Specifically, we use GPT-3.5-Turbo to provide a "Yes-or-No" decision on the correctness of answers and a rating scaled from 0 to 5. We report the percentage of "Yes" responses as Accuracy and the average rating as Score.

\par We conduct all evaluations in a zero-shot way while avoiding the use of elaborate prompts. Besides, we follow the original setup of the models to be reproduced regarding the (maximum) number of frames, frame sampling rate, etc. they applied, ensuring fair and consistent comparisons across all models and benchmarks.

\subsubsection{Major Performance}

\par We compare \ours with state-of-the-art multimodal proprietary models such as Gemini 1.5 Pro~\citep{reid2024gemini}, GPT 4V~\citep{GPT4VisionSystemCard}, and GPT 4o~\citep{HelloGPT4o}, and a series of competitive open-source multimodal models such as VITA~\citep{fu2024vita}, Qwen2-VL~\citep{teamQwen2VLSeeWorld2024}, AnyGPT~\citep{zhan2024anygptunifiedmultimodalllm}, VideoLLaMA 2~\citep{cheng2024videollama}, VideoChat2~\citep{li2024mvbench}, LLaVA-NeXT-Video~\citep{zhang2024llavanext-video}, and Video-LLaVA~\citep{lin2023videollava}. We list major results on general video understanding benchmarks in \autoref{tab:video-perf-1} and results on open-ended video question answering in \autoref{tab:video-perf-2}.

\begin{table}[!ht]
\caption{\textbf{Major results on general VQA benchmarks: MVBench, Egoschema, VideoMME and Perception-Test.} \texttt{max}: Maximum number of sampling frames. $*$: Officially reported results. $\diamondsuit$: Retrieved results from official leaderboard or recent papers. The rest unlabeled results are reproduced by ourselves.}
\label{tab:video-perf-1}
\centering
\begin{tabular}{@{}cccccc@{}}
\toprule
\multicolumn{1}{c|}{}                  & \multicolumn{1}{c|}{}                & \multicolumn{4}{c}{\textbf{General VQA}} \\ \cmidrule(l){3-6} 
\multicolumn{1}{c|}{\multirow{-2}{*}{\textbf{Model}}} &
  \multicolumn{1}{c|}{\multirow{-2}{*}{\textbf{\# Frames}}} &
  \begin{tabular}[c]{@{}c@{}}MVBench\\ (Acc.)\end{tabular} &
  \begin{tabular}[c]{@{}c@{}}Egoschema\\ (Acc.)\end{tabular} &
  \begin{tabular}[c]{@{}c@{}}VideoMME\\ (Acc.)\end{tabular} &
  \begin{tabular}[c]{@{}c@{}}Perception-Test\\ (Acc.)\end{tabular} \\ \midrule
\multicolumn{6}{c}{\cellcolor[HTML]{EFEFEF}\textit{Proprietary Models}}                                                  \\ \midrule
\multicolumn{1}{c|}{Gemini 1.5 Pro}    & \multicolumn{1}{c|}{-}               & 81.3$^\diamondsuit$     & 63.2$^*$     & 75.0$^\diamondsuit$     & -       \\
\multicolumn{1}{c|}{GPT 4o}            & \multicolumn{1}{c|}{-}               & -     & 77.2$^*$     & 71.9$^\diamondsuit$     & -       \\
\multicolumn{1}{c|}{GPT 4V}            & \multicolumn{1}{c|}{-}               & 43.7$^\diamondsuit$     & 55.6$^*$     & 59.9$^\diamondsuit$     & -       \\ \midrule
\multicolumn{6}{c}{\cellcolor[HTML]{EFEFEF}\textit{Open-source Models (Vision-language)}}                                                  \\ \midrule
\multicolumn{1}{c|}{Qwen2 VL (7B)}     & \multicolumn{1}{c|}{2 fps (\texttt{max} 768)} & 67.0$^*$ | 64.4    & 66.7$^*$ | 66.6      & 63.3$^*$ | 59.0      & 62.3$^*$ | 60.3     \\
\multicolumn{1}{c|}{AnyGPT (8B)}  & \multicolumn{1}{c|}{48}               & 33.2     &  32.1    &  29.8    &  29.1   \\
\multicolumn{1}{c|}{VideoLLaMA 2 (7B)} & \multicolumn{1}{c|}{16}              & 54.6$^*$     & 51.7$^*$     & 46.6$^*$     & 51.4$^*$    \\
\multicolumn{1}{c|}{VideoChat2 (7B)}   & \multicolumn{1}{c|}{16}              & 51.1$^*$     & 42.1$^\diamondsuit$     & 33.7$^\diamondsuit$     & 47.3$^\diamondsuit$    \\
\multicolumn{1}{c|}{LLaVA-NeXT-Video (7B)}  & \multicolumn{1}{c|}{32}               & 46.5$^\diamondsuit$     & 43.9$^\diamondsuit$     & 33.7$^\diamondsuit$     & 48.8$^\diamondsuit$    \\
\multicolumn{1}{c|}{Video-LLaVA (7B)}  & \multicolumn{1}{c|}{8}               & 41.0$^\diamondsuit$     & 38.4$^\diamondsuit$     & 39.9$^\diamondsuit$     & 44.3$^\diamondsuit$    \\
\midrule
\multicolumn{6}{c}{\cellcolor[HTML]{EFEFEF}\textit{Open-source Models (Omni-modal)}}          \\ \midrule
\multicolumn{1}{c|}{VITA (8x7B)}       & \multicolumn{1}{c|}{1 fps (\texttt{max} 32)}  & 53.4     & 53.9     & 56.1     & 56.2    \\
\multicolumn{1}{c|}{\textbf{\ours (7B)}}         & \multicolumn{1}{c|}{1 fps (\texttt{max} 48)}  & \textbf{60.9}     & \textbf{58.8}     & \textbf{58.2}     & \textbf{56.8}    \\ \bottomrule
\end{tabular}%
\end{table}

\par \textbf{Results on general video understanding benchmarks.} As shown in \autoref{tab:video-perf-1}, \ours demonstrates competitive results over proprietary models on benchmarks like Egoschema and MVBench, and achieves strong performance across open-source multimodal models, which shows comprehensive video understanding capabilities of \ours.

\par Compared to VITA, a MoE omni-modal LLM with about 12B activated parameters, \ours (7B) outperforms it on all General Video QA benchmarks, and achieve an average improvement of about 4\%. Additionally, \ours excels a series of open-source models such as VideoLLaMA 2, VideoChat2, LLaVA-NeXT-Vide, and Video-LLaVA. Notably, \ours also outperforms the proprietary model GPT 4V on MVBench (43.7\%) and Egoschema (55.6\%).

\par \textbf{Results on open-ended video question answering benchmarks.} The performance on Open-ended VQA is listed in \autoref{tab:video-perf-2}. \ours demonstrates SoTA performance (both Accuracy and Score) on ActivityNet-QA and MSVD-QA across all open-source models, such as the most recent SoTA multimodal models VITA and Qwen2 VL, and outperforms the proprietary model Gemini 1.5 Pro (56.7\%) on ActivityNet-QA. The superior results indicate that \ours is also effective in open-ended question answering, i.e., \ours is more capable of generating informative and descriptive responses.

\begin{table}[!ht]
\caption{\textbf{Major results on ActivityNet-QA and MSVD-QA.} \texttt{max}: Maximum number of sampling frames. $*$: Officially reported results. The rest unlabeled results are reproduced by ourselves.}
\label{tab:video-perf-2}
\centering
\begin{tabular}{@{}cccccc@{}}
\toprule
\multicolumn{1}{c|}{}                      & \multicolumn{1}{c|}{}                & \multicolumn{4}{c}{\textbf{Open-ended VQA}} \\ \cmidrule(l){3-6} 
\multicolumn{1}{c|}{} &
  \multicolumn{1}{c|}{} &
  \multicolumn{2}{c}{ActivityNet-QA} &
  \multicolumn{2}{c}{MSVD-QA} \\
\multicolumn{1}{c|}{\multirow{-3}{*}{\textbf{Model}}} &
  \multicolumn{1}{c|}{\multirow{-3}{*}{\textbf{\# Frames}}} &
  (Acc.) &
  (Score) &
  (Acc.) &
  (Score) \\ \midrule
\multicolumn{6}{c}{\cellcolor[HTML]{EFEFEF}\textit{Proprietary Models}}                                                   \\ \midrule
\multicolumn{1}{c|}{Gemini 1.5 Pro}        & \multicolumn{1}{c|}{-}               & 56.7$^*$     & -       & -       & -      \\
\multicolumn{1}{c|}{GPT 4o}                & \multicolumn{1}{c|}{-}               & 61.9$^*$     & -       & -       & -      \\
\multicolumn{1}{c|}{GPT 4V}                & \multicolumn{1}{c|}{-}               & 59.5$^*$     & -       & -       & -      \\ \midrule
\multicolumn{6}{c}{\cellcolor[HTML]{EFEFEF}\textit{Open-source Models (Vision-language)}}                                                   \\ \midrule
\multicolumn{1}{c|}{Qwen2 VL (7B)}         & \multicolumn{1}{c|}{2 fps (\texttt{max} 768)} & 17.4     & 1.9     & 61.1    & 3.5    \\
\multicolumn{1}{c|}{VideoLLaMA 2 (7B)}     & \multicolumn{1}{c|}{16}              & 50.2$^*$     & 3.3$^*$     & 70.9$^*$    & 3.8$^*$    \\
\multicolumn{1}{c|}{VideoChat2 (7B)}       & \multicolumn{1}{c|}{16}              & 49.1$^*$     & 3.3$^*$     & 70.0$^*$    & 3.9$^*$    \\
\multicolumn{1}{c|}{LLaVA-NeXT-Video (7B)} & \multicolumn{1}{c|}{32}              & 53.5$^*$     & 3.2$^*$     & 67.4    & 3.4    \\
\multicolumn{1}{c|}{Video-LLaVA (7B)}      & \multicolumn{1}{c|}{8}               & 45.3$^*$     & 3.3$^*$     & 70.7$^*$    & 3.9$^*$    \\
\midrule
\multicolumn{6}{c}{\cellcolor[HTML]{EFEFEF}\textit{Open-source Models (Omni-modal)}}          \\ \midrule
\multicolumn{1}{c|}{VITA (8x7B)}           & \multicolumn{1}{c|}{1 fps (\texttt{max} 32)}  & 55.0     & 3.5     & 63.9    & 3.7    \\
\multicolumn{1}{c|}{\textbf{\ours (7B)}}            & \multicolumn{1}{c|}{1 fps (\texttt{max} 48)}  &\textbf{58.6} &\textbf{3.7} &\textbf{72.2} &\textbf{4.0} \\ \bottomrule
\end{tabular}
\end{table}

\subsection{Audio Understanding}
\subsubsection{Evaluation Benchmarks} 
\par To validate the audio understanding capacity of \ours, we present the evaluating results on benchmarks with three tasks: 
\begin{itemize}[leftmargin=*]
    \item \textbf{Automatic Speech Recognition (ASR).} This is a fundamental task for audio-language model pre-training which directly transcribes the audio into the text. For ASR evaluation in the general scene, we report results on the Fleurs~\citep{fleurs2022arxiv} Chinese (\textit{zh}) and English (\textit{en}) test sets, as well as the WenetSpeech~\citep{zhang2022wenetspeech10000hoursmultidomain} \textit{test\_net} dataset. To assess performance in more challenging ASR scenarios,  we include results from the WenetSpeech~\citep{zhang2022wenetspeech10000hoursmultidomain} \textit{test\_meeting} dataset and the KeSpeech~\citep{tang2021kespeech} test set, which evaluate the model's ASR capabilities in `Meeting' and `Chinese dialect' contexts. For WenetSpeech, we use both Word Error Rate (WER) and Character Error Rate (CER) as evaluation metrics, while for others, only WER is used.
    \item \textbf{Speech-to-Text Translation (S2TT).} The task aims to translate the audio signal in the source to the target language. We evaluate the model's S2TT performance between Chinese and English using the zh2en and en2zh subsets of the Covost2~\citep{wang2020covost2massivelymultilingual} dataset, with BLEU~\citep{papineni-etal-2002-bleu} scores as the evaluation metric.
    \item \textbf{AIR-Bench.} The goal of this benchmark is to evaluate the chat capabilities to follow instructions of the given audio. We evaluate chat performance on the chat benchmark~\citep{yang-etal-2024-air} (test set), using Score as the metric.
\end{itemize}

\subsubsection{Major Performance} 
\ours is compared with the state-of-the-art baselines across ASR, S2TT and SER tasks, including the recent leading large audio-language model Qwen2-Audio-Instruct~\cite{chu2024qwen2} and the large omni-modal language model VITA~\cite{fu2024vita}. On top of that, the performance of the classical pre-trained audio language model, Whisper-large-v3~\cite{radford2022robustspeechrecognitionlargescale}, is presented for ASR and the performance of SALMONN~\cite{tang2024salmonn} is presented for S2TT. 

\par \textbf{Results on ASR benchmarks.} \ours exhibits a strong audio transcription capacity in~\autoref{tab:audio-general-asr}. \ours primarily targets the Chinese corpus. In the general Chinese ASR scene, \ours has a 2.0\% WER (2.6\% CER) superiority on the Fleurs test-zh subset and 4.1\% WER (4.2\% CER) improvement on the WenetSpeech test\_net when comparing with Qwen2-Audio-Instruct. The evaluation results on WenetSpeech consistently demonstrate the superiority of our model over VITA. \ours achieves nearly a 50\% improvement in the CER performance of VITA, both in test\_net (\textbf{7.1\%} v.s 12.2\%) and test\_meeting (\textbf{8.9\%} v.s 16.5\%) subsets. On the more challenging Chinese dialect benchmark, KeSpeech, our model maintains a comprehensive lead, with an average CER of 6.7\% over the performance of all dialects. 
Notably, while our model excels in Chinese audio transcription, \ours also maintains robust general ASR performance in English. We achieve 4.7\% of WER, which exceeds Qwen2-Audio-Instruct by 11\% WER.

\begin{table}[!ht]  
\centering  
\caption{\textbf{Major results on} \textbf{Fleurs}, \textbf{WenetSpeech}, \textbf{and} \textbf{KeSpeech}. Test sets of WenetSpeech are evaluated with WER and CER, while other test sets are evaluated only with WER. VITA's evaluation results are officially reported in their paper~\cite{fu2024vita}, marked with $*$. The rest unlabeled results are reproduced by ourselves, and any performance divergence may be attributed to differences in decoding parameters.}  
\begin{tabular}{lccc}  
\toprule  
\multicolumn{1}{c}{\multirow{2}{*}{\textbf{Scene}}} & \multirow{2}{*}{\textbf{Dataset}} & \multirow{2}{*}{\textbf{Model}} & \multirow{2}{*}{\begin{tabular}[c]{@{}c@{}}\textbf{Results} \\ WER (CER) $\downarrow$\end{tabular}} \\
\multicolumn{1}{c}{} & & & \\   
\midrule  
\multirow{7}{*}{General} & \multirow{3}{*}{\begin{tabular}[c]{@{}c@{}}Fleurs \\ \textit{test-zh} | \textit{test-en}\end{tabular}} & Whisper-large-v3 (1.55B) & 12.4 | 7.2 \\
& & Qwen2-Audio-Instruct (7B) & 9.0 | 15.7 \\
& & \ours (7B) & \textbf{7.0 | 4.7} \\   
\cmidrule(l){2-4}   
& \multirow{4}{*}{\begin{tabular}[c]{@{}c@{}}WenetSpeech \\ \textit{test\_net}\end{tabular}} & Whisper-large-v3 (1.55B) & 17.5 (18.5) \\
& & Qwen2-Audio-Instruct (7B) & 11.0 (11.3) \\
& & VITA (8x7B) & - (12.2$^*$) \\
& & \ours (7B) & \textbf{6.9 (7.1)} \\   
\midrule  
\multirow{4}{*}{Meeting} & \multirow{4}{*}{\begin{tabular}[c]{@{}c@{}}WenetSpeech \\ \textit{test\_meeting}\end{tabular}} & Whisper-large-v3 (1.55B) & 30.8 (31.7) \\
& & Qwen2-Audio-Instruct (7B) & 10.7 (10.8) \\
& & VITA (8x7B) & - (16.5$^*$) \\
& & \ours (7B) & \textbf{8.4 (8.9)} \\   
\midrule
\multirow{8}{*}{\begin{tabular}[c]{@{}l@{}}Chinese \\ Dialect\end{tabular}} & \multirow{7}{*}{\begin{tabular}[c]{@{}c@{}}KeSpeech\\ \textit{mandarin} | \textit{beijing} | \textit{southwest}\\ \textit{lan-yin} | \textit{zhongyuan} | \textit{northeast}\\ \textit{jiang-huai} | \textit{ji-lu} | \textit{jiao-liao}\end{tabular}} & Whisper-large-v3 (1.55B) & \begin{tabular}[c]{@{}c@{}}18.7 | 44.8 | 52.9\\ 54.8 | 50.1 | 22.9\\ 54.7 | 47.0 | 50.4\end{tabular} \\
\cmidrule(l){3-4}  
& & Qwen2-Audio-Instruct (7B) & \begin{tabular}[c]{@{}c@{}}5.8 | 9.7 | 10.5\\ 11.0 | 8.2 | 8.4\\ 13.8 | 10.3 | 11.2\end{tabular} \\
\cmidrule(l){3-4}  
& & \ours (7B) & \textbf{\begin{tabular}[c]{@{}c@{}}2.8 | 6.4 | 7.0\\ 7.7 | 6.1 | 5.8\\ 9.0 | 8.3 | 7.2\end{tabular}} \\  
\bottomrule   
\end{tabular}  
\label{tab:audio-general-asr}  
\end{table}

\par \textbf{Results on S2TT and AIR-Bench benchmarks.} In addition to ASR, \ours excels in both S2TT and SER tasks. The evaluation results are summarized in~\autoref{tab:audio-general-s2tt-ser}. Notably, when translating from English to Chinese on the Covost-2 en2zh test set, \ours outperforms Qwen2-Audio-Instruct by approximately 7 BLEU points. For the reverse translation, from Chinese to English, our model's performance on the Covost-2 zh2en test set is comparable to that of Qwen2-Audio-Instruct. On the AirBench, \ours scores 7.42 and 7.26 for speech and sound, respectively, outperforming Qwen2-Audio-Instruct and showcasing \ours's superior ability to generate realistic human speeches and sounds.

\begin{table}[!ht]
    \centering
    \caption{\textbf{Major results on} \textbf{Covost2} \textbf{and} \textbf{AirBench}. $\diamondsuit$ represents the results from the official leaderboard or recent papers. The rest unlabeled results are reproduced by ourselves, and any performance divergence may be attributed to differences in decoding parameters.}
    \small
    \begin{tabular}{@{}lcccc@{}}
        \toprule
        \multicolumn{1}{c}{\multirow{2}{*}{\textbf{Task}}} & 
        \multirow{2}{*}{\textbf{Dataset}} & 
        \multirow{2}{*}{\textbf{Model}} & 
        \multirow{2}{*}{\textbf{Metrics}} & 
        \multirow{2}{*}{\textbf{Results}} \\
        \multicolumn{1}{c}{} & & & & \\ 
        \midrule
        \multirow{3}{*}{S2TT} & 
        \multirow{3}{*}{\begin{tabular}[c]{@{}c@{}}Covost-2\\ \textit{zh2en} | \textit{en2zh}\end{tabular}} & 
        SALMONN (7B) & 
        \multirow{3}{*}{BLEU $\downarrow$} & - | 33.1$^\diamondsuit$ \\
        & & Qwen2-Audio-Instruct (7B) & & \textbf{23.3} | 34.1 \\
        & & \ours (7B) & & 22.1 | \textbf{40.2} \\ 
        \midrule
        \multirow{3}{*}{AIR-Bench} & 
        \multirow{3}{*}{\begin{tabular}[c]{@{}c@{}}Chat Benchmark\\ \textit{speech} | \textit{sound} | \textit{music} | \textit{mix-audio}\end{tabular}} &
        Qwen2-Audio-Instruct (7B) & 
        \multirow{3}{*}{Score $\uparrow$} & 7.18 | 6.99 | \textbf{6.79} | \textbf{6.77} \\
        & & VITA (8x7B) & & 6.40 | 6.59 | 6.59 | 5.94\\ 
        & & \ours (7B) & &\textbf{7.42} | \textbf{7.26} | 6.12 | 5.76\\ 
        \bottomrule
    \end{tabular}
    \label{tab:audio-general-s2tt-ser}
\end{table}

\subsection{Ablation Study}\label{sec: abla_study}

\subsubsection{Image-Language Branch} 

\textbf{Visual encoder.}\quad To compare the performance of different visual encoders in \ours, we conducted experiments across various vision encoders with differing parameter sizes, input resolutions, and output token counts.
We selected five mainstream vision encoders: OpenAI's CLIP series~\citep{radford2021clip}, Google's Siglip series~\citep{zhai2023sigmoid}, Apple's DFN series~\citep{fang2023dfn}, OpenGVLab's InternViT series~\citep{chen2024far}, and BAAI's EVA series~\citep{EVA-CLIP-18B}, totaling 14 models. All models are trained with contrastive learning, with parameters ranging from 300M (ViT-L) to 18B. The training data used during the pre-training of the visual encoders varied from 400M to 10B, with input resolutions spanning from 224$\times$224 to 448$\times$448 and output token counts from 256 to 1024.
All comparative experiments are conducted under the same experimental conditions, specifically using a batch size of 8 and the same data for IFT training (with a data ratio of \textit{Caption: Interleaved data: Pure text} set at \textit{0.45: 0.45: 0.1}).

\begin{table}[!ht]
    \centering
    \caption{\textbf{Comparative study across various vision encoders with differing parameter sizes and input resolutions.} We evaluated the model on 10 benchmarks, including SEEDBench2~\cite{li2023seed}, TextCaps (val)~\cite{sidorov2020textcaps}, TextVQA (val)~\cite{singh2019textvqa}, OCRBench~\cite{liu2024ocrbench}, OCRBench (CN), OKVQA~\cite{marino2019okvqa}, Nocaps~\cite{agrawal2019nocaps}, VQAv2~\cite{goyal2017vqav2}, DocVQA (val)~\cite{mathew2021docvqa}, and GQA~\cite{hudson2019gqa}. We compiled the Average Performance of the model across these 10 benchmarks. Additionally, based on the specific tasks targeted by these benchmarks or their subcategories, we calculated the average scores of the model in six areas: OCR, Nature Image Understanding (NIU), Spatial, Chart, Common Sense Knowledge, and Video.}
    \label{tab:encoder}

    \begin{center}
        \renewcommand{\arraystretch}{1.25}
        \setlength{\tabcolsep}{2.5mm}{
        \resizebox{1.0\linewidth}{!}{
        \begin{tabular}{@{}c|cc|cccccccc@{}}
            \toprule
            \multirow{2}{*}{\textbf{Model}}  & \multirow{2}{*}{\shortstack{\textbf{Params.}}} & \multirow{2}{*}{\shortstack{\textbf{Resolution}}}   & \multirow{2}{*}{\textbf{OCR}} & \multirow{2}{*}{\textbf{NIU}} & \multirow{2}{*}{\textbf{Spatial}}  & \multirow{2}{*}{\textbf{Chart}} & \multirow{2}{*}{\shortstack{\textbf{Common Sense}}} & \multirow{2}{*}{\textbf{Video}} & \multirow{2}{*}{\shortstack{\textbf{Avg.}}} \\ &  &  &  &  &  &  &  &  &  \\
            \midrule
            siglip-so400m-patch14-384  & 428M  & 384 px  & \textbf{44.67} & \underline{56.91} & \underline{41.70} & \textbf{25.00} & \textbf{51.57} & \textbf{40.63} & \textbf{43.80} \\
            clip-vit-large-patch14-336 & 304M  & 336 px  & 30.19 & 55.60 & 41.21 & 15.63 & 48.05 & 37.50 & 39.51 \\
            dfn5b-clip-ViT-H-14-378    & 631M  & 378 px  & 29.05 & 54.75 & 37.50 & 21.88 & 49.22 & 34.38 & 39.14 \\
            \midrule
            InternViT-6B-224px         & 5.9B  & 224 px  & 14.17 & 40.60 & 29.98 & 15.63 & 40.63 & 34.38 & 29.99 \\
            InternViT-6B-448px-v1-5    & 5.5B  & 448 px  & 17.49 & 46.97 & 35.06 & 18.75 & 41.02 & 31.25 & 32.80 \\
            \midrule
            eva-clip-8b                & 7.5B  & 224 px  & 28.86 & 56.61 & 40.92 & \textbf{25.00} & 49.22 & \textbf{40.63} & 41.51 \\
            eva-clip-8b-448            & 7.5B  & 448 px  & \underline{32.66} & \textbf{58.09} & \textbf{43.26} & 21.88 & \underline{49.61} & 37.50 & \underline{41.86} \\
            \bottomrule
        \end{tabular}}
        }
    \end{center}
\end{table}

As shown in \autoref{tab:encoder}, while increasing the resolution does lead to performance improvements (eva-448 vs eva-224, InternViT-6B-224px vs InternViT-6B-448px), the number of encoder parameters does not exhibit a direct relationship with the metrics. Overall, siglip-so400m-patch14-384~\citep{zhai2023sigmoid} achieved the highest average score and excelled in four out of the six tasks, particularly demonstrating outstanding performance in OCR. Considering these factors along with efficiency issues, we ultimately selected siglip-so400m-patch14-384 as the visual encoder for our \ours.

we further explored the impact of using AnyRes~\cite{liu2024llavanext} on the model's visual-language performance. We found that using AnyRes~\cite{liu2024llavanext} results in a significant performance improvement compared to a fixed input of 384 pixels, particularly in tasks that rely on image details, such as visual document understanding, as shown in \autoref{tab:anyres}.

\begin{table}[ht]
    \centering
    \caption{\textbf{Comparative study on AnyRes.} Using AnyRes results in a significant improvement in visual document understanding and OCR.}
    
    \label{tab:anyres}
    
    \begin{center}
        \renewcommand{\arraystretch}{1.1}
        \setlength{\tabcolsep}{1.5mm}{
        \fontsize{9pt}{10pt}\selectfont{
        \begin{tabular}{c|cccc}
            \toprule
            \textbf{Method}          & \textbf{TextVQA (val)}& \textbf{DocVQA (val)}& \textbf{InfographicVQA}& \textbf{OCRBench}\\
            \midrule
            Baseline& 66.48& 72.61& 47.54& 76.92\\
            
            Baseline+Anyres& 69.13& 87.48& 62.80& 78.44\\
            
            \bottomrule
        \end{tabular}}}
    \end{center}
    \vspace{-2mm}
\end{table}

\textbf{Projector}.\quad Regarding the projector, we compared the following approaches:
\textbf{(1) MLP}: Directly passes through a two-layer MLP, aligning the dimensions to that of the LLM, without reducing the number of image tokens.
\textbf{(2) C-abs}: Passes through two convolutional layers and one pooling layer, aligning the dimensions to that of the LLM while reducing the number of tokens as needed (e.g., from 576 to 144).
\textbf{(3) Concat}: Concatenates adjacent tokens and then processes them through a MLP, allowing for token reduction but increasing the number of parameters (as the MLP's input dimension increases).
\textbf{(4) Mean Pool}: Applies a convolutional layer with a stride of 2 for pooling before passing the tokens through a two-layer MLP, enabling token reduction while maintaining a consistent parameter count with the MLP.

In early experiments, we found that models trained with different projectors exhibited little difference in general image understanding, but shows disparities in Chinese OCR comprehension after adding 1 million pure Chinese OCR VQA data.
The results showed that while the model with the C-abs projector struggled to learn Chinese OCR capabilities, the model with the MLP projector began fitting the data and demonstrated zero-shot capability after 0.75 epochs.
Ultimately, we ranked the projectors as follows: MLP > Mean Pool > Concat > C-abs. 
On the other hand, to minimize the number of tokens per sub-image after the AnyRes operation (where MLP produces 729 tokens, while Mean Pool, Concat, and C-abs each produce 182 tokens), we chose Mean Pool as our visual projector.

\subsubsection{Video-Language Branch}

\par For video modality, we conduct an ablation study from three perspectives to thoroughly investigate the impact of various factors on model performance. 
\par \textbf{Number of frames.}\quad Within the constraints of the context length, we systematically adjust the frame sampling rate to control the maximum number of input video frames.

\par \textbf{Resolution of vision encoder}.\quad We explore the effect of different vision encoder resolutions on the model's ability to extract meaningful visual features. Our investigation spans from fixed resolutions (such as 384 $\times$ 384 pixels) to dynamic resolution approaches like AnyRes.

\par \textbf{Video-language pre-training.}\quad We evaluate the model's performance both with and without video-language pre-training. This comparison helps us quantify the benefits of leveraging large-scale multimodal datasets for pre-training, potentially enhancing the model's ability to understand video-text relationships and generalize across various video-understanding tasks.

\begin{table}[!ht]
\caption{\textbf{Ablation study on Video-language branch.} We analyze the influence of number of frames, resolution of vision encoder, and video-language pre-training.}
\label{tab:ablation-video}
\centering
\begin{tabular}{@{}cccc|cccc@{}}
\toprule
\textbf{w/ Pre-training} & \textbf{Resolution} & \textbf{\# Frames} & \textbf{\# Tokens} & \textbf{MVBench} & \textbf{VideoMME} & \textbf{ActivityNet-QA} & \textbf{Avg.} \\ \midrule
$\checkmark$  & 384 px   & 64 & 45 & 50.5 & 56.9 & 56.8 & 54.7 \\
$\checkmark$  & 384 px   & 48 & 45 & 47.6 & 54.6 & 48.1 & 50.1 \\
$\times$ & 384 px   & 64 & 45 & 46.8 & 56.0 & 44.6 & 49.1 \\
$\checkmark$ & AnyRes & 48 & 182 - 546 & 60.9 & 58.2 & 58.6 & 59.2 \\ \bottomrule
\end{tabular}%
\end{table}

\par As demonstrated in \autoref{tab:ablation-video}, the model's performance in video comprehension is significantly impacted by the number of input frames processed. When the quantity of input frames is decreased from 64 to 48, there is a notable decline (from 54.7\% to 50.1\% average) in the model's ability to understand and interpret video content.

\par When testing the model by inputting a total of 48 frames, it has been observed that the version of the model that utilizes AnyRes technology demonstrates superior performance compared to the version that operates with a fixed resolution set at 384$\times$384. This performance advantage is evident across various benchmarks, including MVBench, VideoMME, and ActivityNet-QA. In fact, the model with AnyRes enabled shows an average improvement of around 5\% over its fixed-resolution counterpart.

\par Besides, from the first and third rows of the table, it shows that incorporating video-text pre-training has a pronounced impact on the model's video understanding capability. For instance, in MVBench, the model without pre-training lags approximately 6\% behind the one with pre-training.

\par Overall, we found that increasing the number of video frames, enhancing the resolution of the visual encoder, and incorporating video-text data during the pre-training stage all contribute to improving the model's capabilities of understanding videos. We will leave the exploration of these three factors in situations where the input exceeds the context length (increasing number of frames and resolution) for future work.

\subsubsection{Audio-Language Branch}\label{sec: audio-ablation}

\par The audio projector in the audio-language branch plays a key role in bridging the representations of audio and natural language modalities. Notably, our newly introduced projector with Conv-GMLP demonstrates the great performance robustness of the feature down-sampling rate. 

\begin{wrapfigure}{r}{.45\textwidth}
    \centering
    \vspace{-1.8em}
    \includegraphics[width=.45\textwidth]{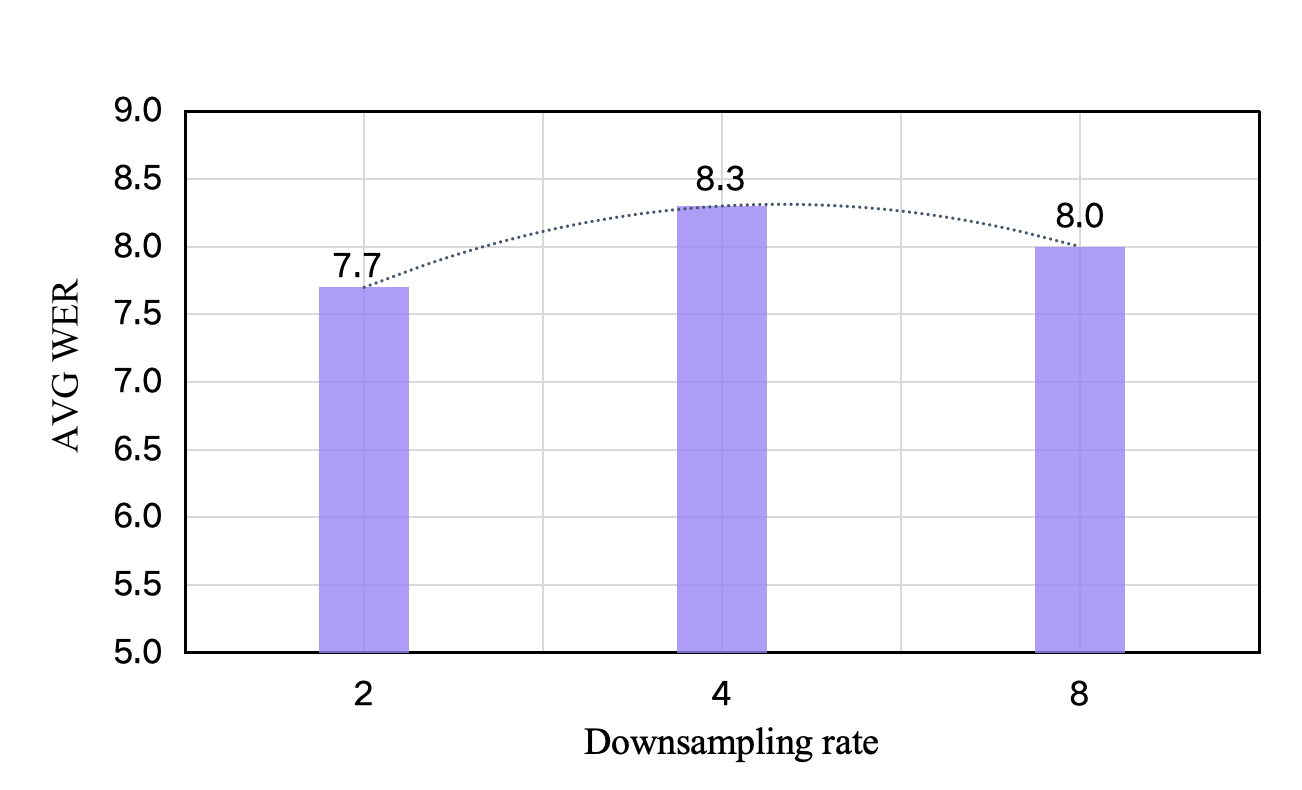}
    \caption{\textbf{Ablation study on down-sampling rate.} Average WER across multiple test sets (Fleurs zh/en, WenetSpeech net/meeting, and KeSpeech) using various down-sampling rates of Conv-GMLP.}
    \label{fig:audio_ablation}
    \vspace{-2em}
\end{wrapfigure}

\par For analysis, we measure the average WER on all our ASR benchmarks across Fleurs, WenetSpeech and KeSpeech by training a 1.5B audio-language model with three different down-sampling rates 2, 4, 8. To simulate the actual training of the audio branch in \ours, we only train our audio encoder and projector while keeping the LLM frozen. This setup is consistent with the configuration described in Section~\ref{sec: audio_branch_training}. 

\par From the \autoref{fig:audio_ablation}, we observe that when the down-sampling rate is set to 2, the audio-language model achieves the best ASR performance, with an average WER of 7.7\%. When the down-sampling rate is adjusted to 4 and 8, there is a slight degradation in ASR performance, but the decrease is minimal (ranging from 0.3\% to 0.6\%). Surprisingly, despite the greater degree of down-sampling, the model with a rate of 8 outperforms the one with a rate of 4 (8.0\% vs. 8.3\%). This highlights the exceptional sequence compression capability of the Conv-GMLP.

\subsubsection{Multimodal Supervised Fine-Tuning}

\autoref{tab:sft_img} and \autoref{tab:sft_video} compare the performance of \ours on various image and video benchmarks with and without multimodal supervised fine-tuning (SFT). The results indicate that the model exhibits superior overall performance after undergoing multimodal SFT compared to the version that only undergoes instruction fine-tuning (IFT).
This improvement can be attributed to the use of high-quality, diverse instructions and our SFT data construction method, which avoid compromising the base model's capabilities. (See Section~\ref{sft_data} for more details.)

\begin{table}[!ht]
\caption{\textbf{Performances of \ours on image tasks before and after the supervised fine-tuning stage.} Generally, the model's performance has improved across most image benchmarks.}
\label{tab:sft_img}
\centering
\resizebox{1.0\linewidth}{!}{
\begin{tabular}{@{}ccccc|ccccc@{}}
\toprule
\multicolumn{1}{c|}{}         & \multicolumn{4}{c}{\textbf{Multi-choice Question}}    & \multicolumn{4}{c}{\textbf{Visual Question Answering}} \\ 
\cmidrule(l){2-9} 
\multicolumn{1}{c|}{}         & MMBench  & MMBench-CN    & MMMU    & SEED-IMG    & ChartQA   & MathVista   & MMVet  & RealWorldQA \\
\multicolumn{1}{c|}{\multirow{-3}{*}{\textbf{Method}}}   & (Acc.)  & (Acc.)      & (Acc.)    & (Acc.)      & (Acc.) & (Acc.) & (Acc.) & (Acc.) \\ 
\midrule
\multicolumn{1}{c|}{IFT}      & 75.6      & 69.3         & \textbf{48.3}    & 73.0        & 76.0      & 51.6        & 55.0   & \textbf{62.9}\\
\multicolumn{1}{c|}{SFT}  & \textbf{76.2}      & \textbf{74.9}         & 47.3    & \textbf{74.1}        & \textbf{79.6}      & \textbf{51.9}        & \textbf{65.4}   & 62.6\\
\bottomrule
\end{tabular}}
\end{table}

\begin{table}[!ht]
\caption{\textbf{Performances on video tasks before and after the supervised fine-tuning stage.} Multimodal supervised fine-tuning brings significant improvements for the vast majority of video benchmarks.}
\label{tab:sft_video}
\centering
    \begin{tabular}{@{}ccccc|ccccc@{}}
    \toprule
    \multicolumn{1}{c|}{}         & \multicolumn{4}{c}{\textbf{General VQA}}    & \multicolumn{4}{c}{\textbf{Open-ended VQA}} \\ 
    \cmidrule(l){2-9}
    \multicolumn{1}{c|}{}         & Egoschema        & MVBench            & VideoMME           & Perception      & \multicolumn{2}{c}{ActivityNet-QA}   & \multicolumn{2}{c}{MSVD-QA} \\
    \multicolumn{1}{c|}{\multirow{-3}{*}{\textbf{Method}}}    & (Acc.)  & (Acc.) & (Acc.) & (Acc.) & (Acc.) & (Score) & (Acc.) & (Score) \\ 
    \midrule
    \multicolumn{1}{c|}{IFT}        & 54.0             & \textbf{61.3}              & 56.3     & \textbf{56.9}   & 55.4           & 3.6                 & 66.6              & 3.8\\
    \multicolumn{1}{c|}{SFT}    & \textbf{58.8}    & 60.9              & \textbf{58.2}     & 56.8            & \textbf{58.6}  & \textbf{3.7}        & \textbf{72.2}     & \textbf{4.0}\\
    \bottomrule
    \end{tabular}
\end{table}

\section{Conclusion}

In this work, we have open-sourced \ours as a step toward developing a truly omni-modal LLM that encompasses all human senses. With omni-modal pretraining and fine-tuning using high-quality omni-modal data, this version of \ours has achieved leading levels in integrating comprehension across video, image, text, and audio.
Despite its promising performance, there remains significant room for improvement in the foundational capabilities across each individual modality. This include (1) enhancing text extraction capabilities, (2) supporting longer video understanding, (3) developing an end-to-end TTS system integrated with LLMs, and (4) improving the ability to comprehend not only human voices but also natural environmental sounds, such as flowing water, bird calls, and collision noises, among others.

We anticipate efforts from both academia and industry in the field, and we believe that the expansion of model modalities, along with simultaneous advancements in enhancing model capabilities, will ultimately bring the dream of Artificial General Intelligence closer to reality.

\bibliography{references}
\bibliographystyle{plain}

\end{document}